\pdfoutput=1

\documentclass[11pt]{article}

\usepackage{acl}

\usepackage{times}
\usepackage{latexsym}

\usepackage[T1]{fontenc}

\usepackage[utf8]{inputenc}

\usepackage{microtype}

\usepackage{inconsolata}

\usepackage{amsmath}
\usepackage{multirow}
\usepackage{bbm}
\usepackage{graphicx}
\usepackage{amssymb}
\usepackage{booktabs}
\usepackage{adjustbox}
\usepackage{xspace}
\usepackage{algpseudocode}
\usepackage{algorithm}
\usepackage{times}
\usepackage{latexsym}
\usepackage{graphicx}
\usepackage{caption}
\usepackage{subcaption}
\usepackage{hyperref}
\usepackage{color, colortbl}
\usepackage{tablefootnote}

\DeclareMathOperator*{\argmaxA}{arg\,max} 

\title{Multilingual Few-Shot Learning via Language Model Retrieval}

\author{Genta Indra Winata, Liang-Kang Huang, Soumya Vadlamannati, Yash Chandarana \\
  Bloomberg \\
  \texttt{\{gwinata, lhuang214, svadlamanna1, ychandarana\}@bloomberg.net}}

\begin{document}
\maketitle
\begin{abstract}
Transformer-based language models have achieved remarkable success in few-shot in-context learning and drawn a lot of research interest. However, these models' performance greatly depends on the choice of the example prompts and also has high variability depending on how samples are chosen. In this paper, we conduct a comprehensive study of retrieving semantically similar few-shot samples and using them as the context, as it helps the model decide the correct label without any gradient update in the multilingual and cross-lingual settings. We evaluate the proposed method on five natural language understanding datasets related to intent detection, question classification, sentiment analysis, and topic classification. The proposed method consistently outperforms random sampling in monolingual and cross-lingual tasks in non-English languages.

\end{abstract}

\section{Introduction}

Transformer-based language models (LMs)~\cite{devlin2019bert,raffel2020exploring,xue2021mt5,lewis2020bart,liu2020multilingual,Radford2019LanguageMA,brown2020language,mesh-transformer-jax,zhang2022opt} have shown strong capability in few short learning with prompts. This capability allows them to adapt quickly to diverse tasks from a small amount of data without tedious tuning, and is particularly useful in low-resource settings where the data of the target task, domain or language is fairly limited~\cite{louvan2020recent,schick2021few,lester2021power,perez2021true,winata2022cross}.
Among various studies that aimed to understand and further improve the few-shot prompt learning capabilities, some recent literature present evidences that shows the model's performance on few-shot learning could vary greatly according to the choice of the prompting examples. However, these studies so far were rather preliminary and limited in scope.  A comprehensive study about the impact and strategies for choosing prompts is yet to be done~\cite{liu2021makes}.
\begin{figure*}[!ht]
	\centering
	\includegraphics[width=0.95\linewidth]{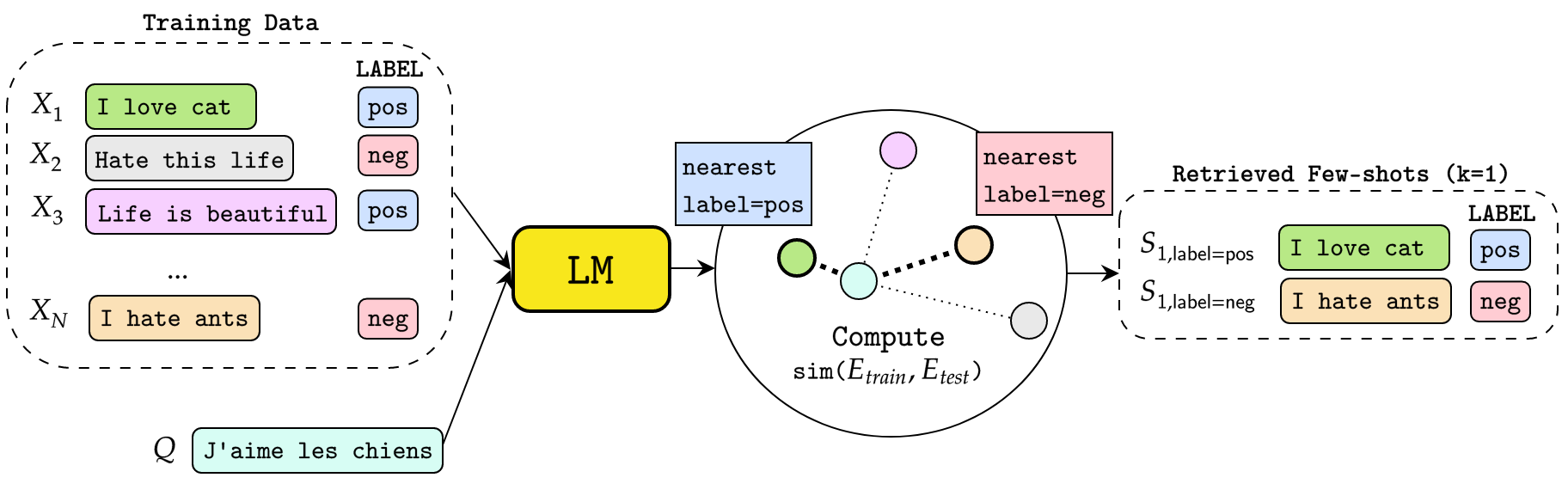}   
	\caption{LM based retrieval. In this example, the context is in \textbf{English} and the query is in \textbf{French}.}
	\label{fig:architecture}
\end{figure*}
\citet{liu2021makes} studied the strategy of choosing semantically similar examples as prompts and showed that examples with similar representations as the queries serve as better prompting examples. However, their work demonstrated this finding solely on English tasks. While the sampling strategy of the in-context learning on non-English and cross-lingual tasks have not been explored, and their work are limited to random sampling~\cite{winata2021language,lin2021few,huang2022zero}. 
Thus, understanding how useful each data sampling strategy can help us to apply cross-lingual transfer more effectively.

In this paper, we conduct a comprehensive study of applying the semantic-based sampling strategy on a wide range of multilingual and cross-lingual tasks and state-of-the-art transformer LMs. From the result,
we show the effectiveness of leveraging semantically similar samples as context and evaluate them in the downstream natural language understanding (NLU) tasks in four different languages: English, French, German, and Spanish. We explore the in-context learning not only on the monolingual setting, but also on the cross-lingual setting, where we retrieve samples from a different language than the language of the given text. 
We find that applying semantically similar few-shot samples on the models consistently outperforms random sampling performance, and retrieving samples from multilingual LM is able to select semantically similar samples across different languages. Furthermore, in most tasks, the performance of the model decreases when we use less similar samples as context showing the importance of samples selection in the multilingual and cross-lingual in-context learning.


\section{Multilingual Language Model Retrieval}

We define and formalize the LM-based retrieval and then describe how we utilize the retrieved few-shot samples in the in-context learning setting.

\subsection{Preliminaries}
Let us define $\mathcal{D}$ as the distribution over the dataset. We propose to utilize a pre-trained multilingual LM for retrieving samples from the training samples. Given a set of training samples $X = \{X_1, X_2, ..., X_N\}$ in source language $L_{1}$ and a test sample $Q$ in target language $L_{2}$, we show the retrieval process in Figure~\ref{fig:architecture}. We compute the $D$-dimensional sentence-level embeddings $E_{Q} \in \mathbb{R}^D$ and training samples $E_{X_i} \in \mathbb{R}^D$ by aggregating the subword embeddings through average pooling over all subwords. We compute the distance $d$ using a distance function $sim(E_{Q}, E_{X_i})$ from the embeddings of the query and the test sample and retrieve top-$k$ nearest samples $S = \{[S_{1,l_1},...,S_{k,l_1}], ..., [S_{1,l_{|L|}},...,S_{k,l_{|L|}}]\}$, where $L = \{l_1,...,l_{|L|}\}$ is the set of all labels and $S_{i,l_j}$ denotes the $i^{th}$ closest sample to $Q$ with ground truth label $l_j$. 

\subsection{In-Context Learning}
Let us define $P$ as the prompt and LM $\theta$. The prompt $P=[I, S, Q]$ is a concatenation of task instruction $I$, few-shot samples $S$, and query $Q$. We pass the prompt $P$ as input to the model $\theta$ and the model computes the probability distribution of predicting the next word $p(l|P)$, where $l \in L$ and $L$ is the set of all labels. Then, we take the highest label probability as the predicted label $\hat{l}$ and it is formulated as follows:
\begin{align}
    \hat{l} &= \argmaxA_{l} p(l|P), \\
    p(l|P) &= \prod_{t=1}^T p(l_t|P,l_{<t}),
\end{align}
where $l_{<t}$ is the previous predicted tokens and label $l$ can be tokenized into $T$ subwords tokens $\{l_1,...,l_t,...,l_T\}$. 

\paragraph{Standardization}
We study the standardizing method on our representation to correct for rogue dimensions by following~\citet{timkey2021all}. 
By post-processing the train and test embeddings, we conjecture that we are able to retrieve few-shot examples whose similarity function with the query aligns more with human similarity judgements. 




\begin{figure*}[!ht]
	\centering
	\includegraphics[width=\linewidth]{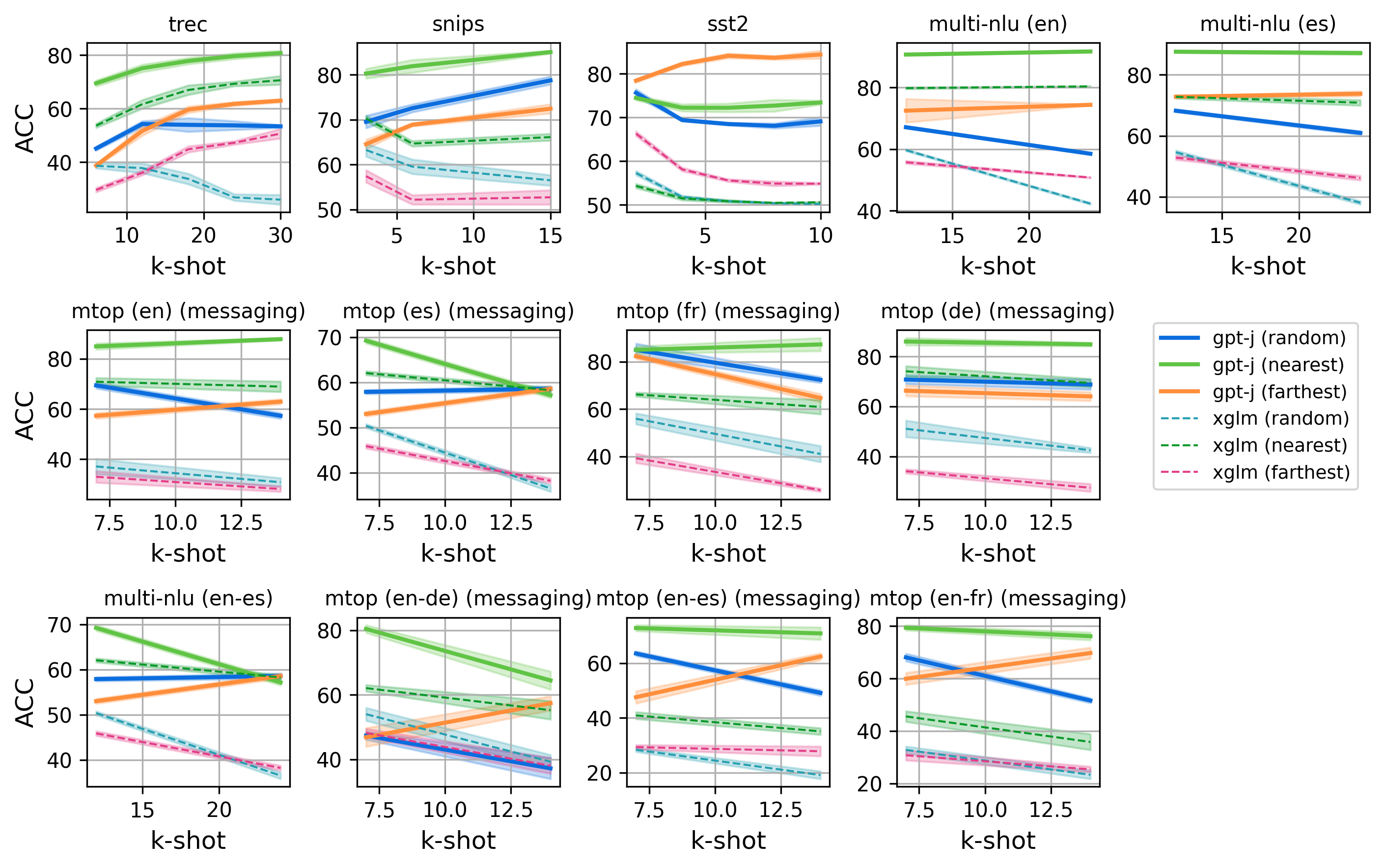}   
	\caption{Few-shot in-context learning results on the monolingual and cross-lingual settings in accuracy over 5 runs. We compare three retrieval strategies (random, nearest, and farthest) on GPT-J NEO and XGLM models.}
	\label{fig:all-results}
\end{figure*}

\section{Experiments}
\subsection{Datasets}
In this work, we experiment the proposed strategy on five datasets with a variety of monolingual and multilingual downstream NLU tasks, \textbf{intent detection:} SNIPS~\cite{coucke2018snips}; Multi-NLU~\cite{schuster-etal-2019-cross-lingual}; MTOP~\cite{li-etal-2021-mtop}; \textbf{topic classification}: TREC~\cite{li-roth-2002-learning}; \textbf{sentiment dataset:} SST2~\cite{wang2018glue};.
We filter the training set if they are seen in the test set since we would to ensure that the test set is not seen in the training set.
While examining the datasets we observe overlaps in all five of them, where identical samples are presented in both training and testing set. We decide to filter those samples from training set to avoid retrieval strategies taking advantage of the fact by selecting prompts identical to the query. The overlap rates for each dataset are shown in Table~\ref{overlap} in the Appendix. We can see that the rate is particularly high in Multi-NLU.

\begin{figure}[!t]
	\centering
	\includegraphics[width=0.89\linewidth]{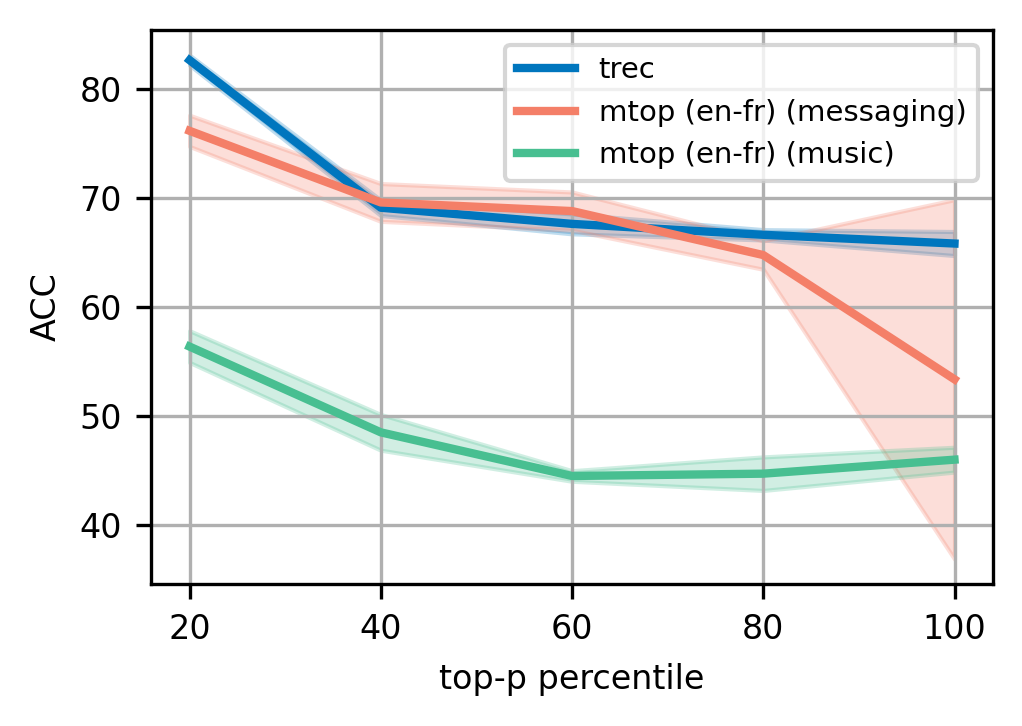}   
	\caption{Results on various sampling intervals over 5 runs in accuracy. The shaded area represents the standard deviation. Lower $p$ means we retrieve samples that are contextually more similar to the query based on similarity measure.}
	\label{fig:interval-results}
\end{figure}

\subsection{Setup}
We perform few-shot learning by prompting an English decoder-only LM, GPT-J NEO \texttt{(1.3GB)}~\cite{mesh-transformer-jax} and a multilingual LM, XGLM \texttt{(1.7B)}~\cite{lin2021few}, and run inference with a single V100 32GB GPU. For retrieving prompts, we compute the sentence-level representation of the query and training samples with XLM-R$_{\text{BASE}}$~\cite{conneau2020unsupervised} respectively, and use euclidean distance and cosine similarity, with and without normalizing the embeddings, to measure the semantic similarity. We compare the effectiveness among three retrieval strategies: \texttt{random}, \texttt{nearest}, and \texttt{farthest}. \texttt{random} means we select the few-shot samples randomly; \texttt{nearest} means we select the samples with the highest similarity scores; and \texttt{farthest} means we select the samples with the lowest similarity scores. Experiments are conducted on both \texttt{monolingual} and \texttt{cross-lingual} settings. The former picks prompts in the same language as the query, while the latter picks prompts of a language different from the query.

\section{Results and Discussion}
\begin{figure}[!t]
	\centering
	\includegraphics[width=0.95\linewidth]{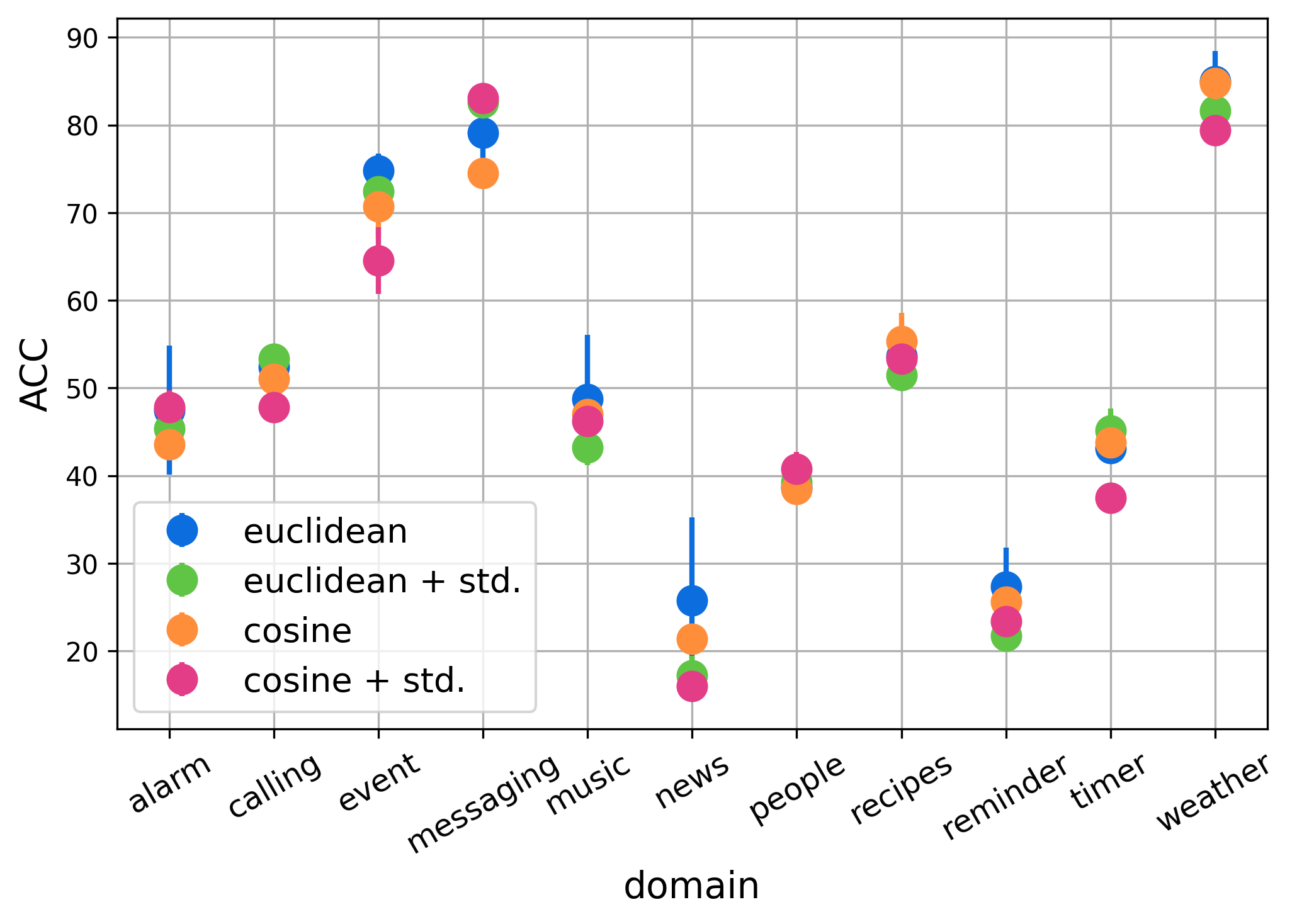}   
	\caption{Results with different similarity measures on MTOP (en-fr). We take 5 samples from each label to form the context.}
	\label{fig:measure-results}
\end{figure}

\begin{figure}[!t]
	\centering
	\includegraphics[width=\linewidth]{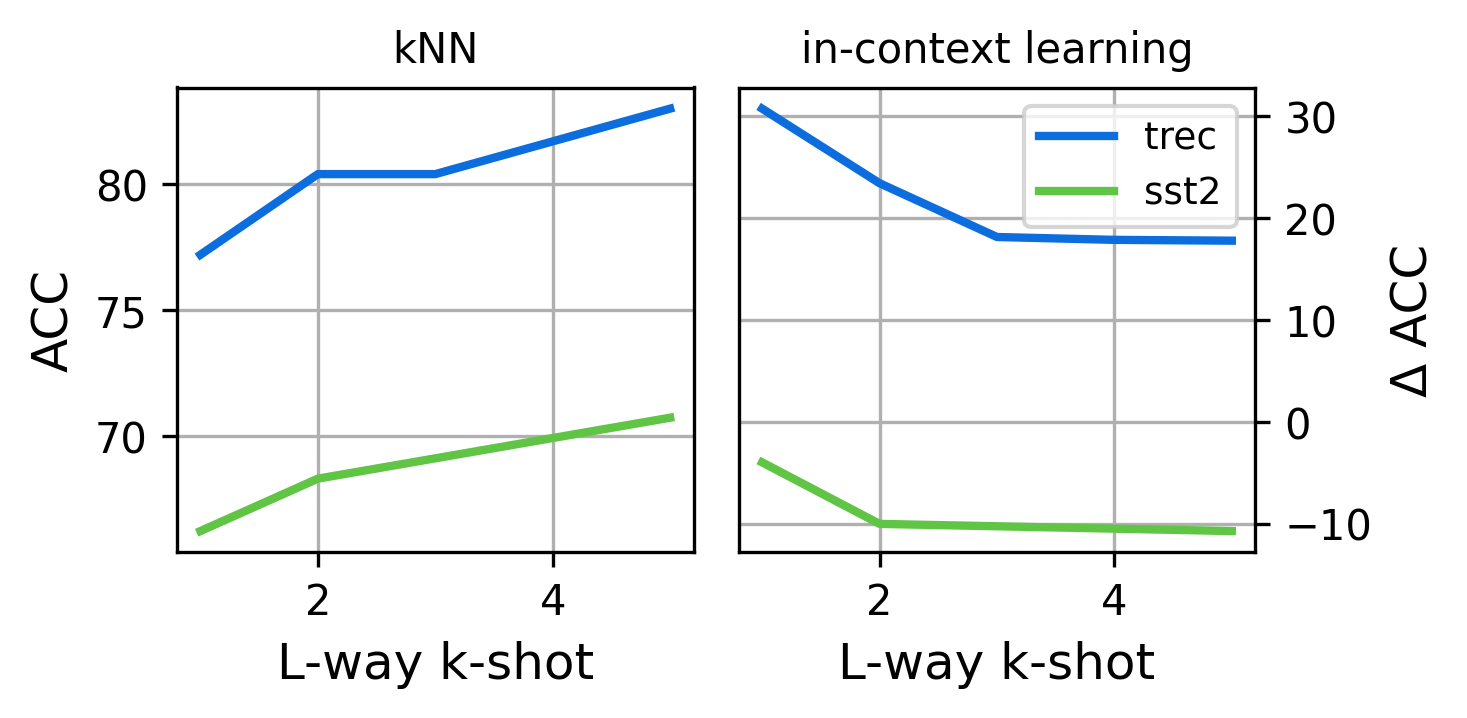}   
	\caption{(a) kNN performance, and (b) Accuracy difference between nearest and farthest retrieval strategies.}
	\label{fig:knn-results}
\end{figure}

We show the in-context learning results on different datasets in Figure~\ref{fig:all-results}. Each plot presents the few-shot learning performance of GPT-J NEO and XGLM with the three retrieval strategies, while varying the value of $k$. In general, the nearest strategy consistently outperforms random and farthest, except in SST2 dataset (this observation is further discussed later in Figure~\ref{fig:knn-results}). Interestingly, the multilingual 1.7B XGLM model does not perform as well as the English 1.3B GPT-J NEO model, under English, non-English and cross-lingual settings.
\paragraph{Sampling From Similarity Score Distribution}
Figure~\ref{fig:interval-results} shows the dynamics of performance on various sampling intervals. The performance of the model decreases when the $p$ is larger on both TREC and MTOP datasets, and the standard deviation is relatively higher when collecting less similar samples.
\paragraph{Similarity Measures} Figure~\ref{fig:measure-results} shows the performance with different similarity measures evaluated on MTOP (en-fr). There is no clear winner for which similarity measure performs the best, and it shows that euclidean distance can perform similarly as cosine distance, and empirically, there is no evidence that applying standardization is useful.

\paragraph{Comparison kNN vs. In-Context Learning}
To further understand the fact that the farthest strategy outperforms nearest with both LMs on the SST2 dataset, Figure~\ref{fig:knn-results} shows the results of the kNN performance and the performance gap of the nearest and farthest retrieval strategies ($\Delta$ ACC). The fact that kNN performs relatively worse on SST2 suggests that in SST2 dataset, semantically similar examples do not necessary have the same labels, which explains why we observe a different performance gap between nearest and farthest on these two datasets. This is either due to the characteristic of the dataset or an indication that the semantic representation produced by XLM-R$_{\text{BASE}}$~\cite{conneau2020unsupervised} is not accurate in SST2.







\section{Related Work}
\label{sec:related-work}

\paragraph{Few-shot In-Context Learning}
Recent work on few-shot in-context learning uses LMs to solve NLP tasks~\cite{petroni2019language,brown2020language,gao2020making,madotto2020language,zhao2021calibrate,schick2021s,lin2021leveraging,utama2021avoiding, srivastava2022beyond}. In this approach, the appropriate prompts were selected to trigger the LMs to behave so that they can predict the desired output~\cite{liu2021pre}. Few-shot learning has been applied to various applications in multilingual and cross-lingual task~\cite{winata2021language,lin2021few,winata2022cross}, zero-shot cross-lingual~\cite{huang2022zero}, and dialogue system ~\cite{madotto2021few}.



\paragraph{Cross-lingual Retrieval}
Multilingual LMs have been utilized for document and sentence-level information retrieval (IR) tasks across a large number of language pairs~\cite{litschko2021evaluating}. The effects of language relatedness within a multilingual IR have been studied by~\citet{chew2008effects}. They found that choosing languages that are more related the target language can boost the precision.


\section{Conclusion}
\label{sec:conclusion}
We apply a simple yet effective LM-based retrieval method to retrieve semantically similar few-shot samples as the context on multilingual and cross-lingual in-context learning. We present the effectiveness of retrieving semantically similar few-shot samples to inform the LM to predict the correct label without any gradient updates. We find that retrieving similar samples consistently outperforms random sampling, when the semantically similar examples have the same labels as query. Furthermore, the performance of the model decreases when we use less similar samples as context showing the importance of samples selection in the multilingual and cross-lingual in-context learning. Furthermore, we conduct further systematic analysis on applying retrieval strategies from different percentile intervals and similarity measures.

\section*{Limitations}
We run our experiments only on GPT-J and XGLM models due to limited resource constraint. We leave the exploration of larger or other monolingual and multilingual variant language models (e.g., BLOOM~\cite{scao2022bloom}) as future work.



\bibliography{anthology,custom}

\begin{thebibliography}{38}
\expandafter\ifx\csname natexlab\endcsname\relax\def\natexlab#1{#1}\fi

\bibitem[{Brown et~al.(2020)Brown, Mann, Ryder, Subbiah, Kaplan, Dhariwal,
  Neelakantan, Shyam, Sastry, Askell et~al.}]{brown2020language}
Tom Brown, Benjamin Mann, Nick Ryder, Melanie Subbiah, Jared~D Kaplan, Prafulla
  Dhariwal, Arvind Neelakantan, Pranav Shyam, Girish Sastry, Amanda Askell,
  et~al. 2020.
\newblock Language models are few-shot learners.
\newblock \emph{Advances in neural information processing systems},
  33:1877--1901.

\bibitem[{Chew and Abdelali(2008)}]{chew2008effects}
Peter~A Chew and Ahmed Abdelali. 2008.
\newblock The effects of language relatedness on multilingual information
  retrieval: a case study with indo-european and semitic languages.
\newblock In \emph{Proceedings of the 2nd workshop on Cross Lingual Information
  Access (CLIA) Addressing the Information Need of Multilingual Societies}.

\bibitem[{Conneau et~al.(2020)Conneau, Khandelwal, Goyal, Chaudhary, Wenzek,
  Guzm{\'a}n, Grave, Ott, Zettlemoyer, and Stoyanov}]{conneau2020unsupervised}
Alexis Conneau, Kartikay Khandelwal, Naman Goyal, Vishrav Chaudhary, Guillaume
  Wenzek, Francisco Guzm{\'a}n, {\'E}douard Grave, Myle Ott, Luke Zettlemoyer,
  and Veselin Stoyanov. 2020.
\newblock Unsupervised cross-lingual representation learning at scale.
\newblock In \emph{Proceedings of the 58th Annual Meeting of the Association
  for Computational Linguistics}, pages 8440--8451.

\bibitem[{Coucke et~al.(2018)Coucke, Saade, Ball, Bluche, Caulier, Leroy,
  Doumouro, Gisselbrecht, Caltagirone, Lavril et~al.}]{coucke2018snips}
Alice Coucke, Alaa Saade, Adrien Ball, Th{\'e}odore Bluche, Alexandre Caulier,
  David Leroy, Cl{\'e}ment Doumouro, Thibault Gisselbrecht, Francesco
  Caltagirone, Thibaut Lavril, et~al. 2018.
\newblock Snips voice platform: an embedded spoken language understanding
  system for private-by-design voice interfaces.
\newblock \emph{arXiv preprint arXiv:1805.10190}, pages 12--16.

\bibitem[{Devlin et~al.(2019)Devlin, Chang, Lee, and
  Toutanova}]{devlin2019bert}
Jacob Devlin, Ming-Wei Chang, Kenton Lee, and Kristina Toutanova. 2019.
\newblock Bert: Pre-training of deep bidirectional transformers for language
  understanding.
\newblock In \emph{Proceedings of the 2019 Conference of the North American
  Chapter of the Association for Computational Linguistics: Human Language
  Technologies, Volume 1 (Long and Short Papers)}, pages 4171--4186.

\bibitem[{Gao et~al.(2020)Gao, Fisch, and Chen}]{gao2020making}
Tianyu Gao, Adam Fisch, and Danqi Chen. 2020.
\newblock Making pre-trained language models better few-shot learners.
\newblock \emph{arXiv preprint arXiv:2012.15723}.

\bibitem[{Huang et~al.(2022)Huang, Ma, Zhang, Wei, and Wang}]{huang2022zero}
Lianzhe Huang, Shuming Ma, Dongdong Zhang, Furu Wei, and Houfeng Wang. 2022.
\newblock Zero-shot cross-lingual transfer of prompt-based tuning with a
  unified multilingual prompt.
\newblock \emph{arXiv preprint arXiv:2202.11451}.

\bibitem[{Lester et~al.(2021)Lester, Al-Rfou, and Constant}]{lester2021power}
Brian Lester, Rami Al-Rfou, and Noah Constant. 2021.
\newblock The power of scale for parameter-efficient prompt tuning.
\newblock In \emph{Proceedings of the 2021 Conference on Empirical Methods in
  Natural Language Processing}, pages 3045--3059.

\bibitem[{Lewis et~al.(2020)Lewis, Liu, Goyal, Ghazvininejad, Mohamed, Levy,
  Stoyanov, and Zettlemoyer}]{lewis2020bart}
Mike Lewis, Yinhan Liu, Naman Goyal, Marjan Ghazvininejad, Abdelrahman Mohamed,
  Omer Levy, Veselin Stoyanov, and Luke Zettlemoyer. 2020.
\newblock Bart: Denoising sequence-to-sequence pre-training for natural
  language generation, translation, and comprehension.
\newblock In \emph{Proceedings of the 58th Annual Meeting of the Association
  for Computational Linguistics}, pages 7871--7880.

\bibitem[{Li et~al.(2021)Li, Arora, Chen, Gupta, Gupta, and
  Mehdad}]{li-etal-2021-mtop}
Haoran Li, Abhinav Arora, Shuohui Chen, Anchit Gupta, Sonal Gupta, and Yashar
  Mehdad. 2021.
\newblock \href {https://doi.org/10.18653/v1/2021.eacl-main.257} {{MTOP}: A
  comprehensive multilingual task-oriented semantic parsing benchmark}.
\newblock In \emph{Proceedings of the 16th Conference of the European Chapter
  of the Association for Computational Linguistics: Main Volume}, pages
  2950--2962, Online. Association for Computational Linguistics.

\bibitem[{Li and Roth(2002)}]{li-roth-2002-learning}
Xin Li and Dan Roth. 2002.
\newblock \href {https://www.aclweb.org/anthology/C02-1150} {Learning question
  classifiers}.
\newblock In \emph{{COLING} 2002: The 19th International Conference on
  Computational Linguistics}.

\bibitem[{Lin et~al.(2021{\natexlab{a}})Lin, Mihaylov, Artetxe, Wang, Chen,
  Simig, Ott, Goyal, Bhosale, Du et~al.}]{lin2021few}
Xi~Victoria Lin, Todor Mihaylov, Mikel Artetxe, Tianlu Wang, Shuohui Chen,
  Daniel Simig, Myle Ott, Naman Goyal, Shruti Bhosale, Jingfei Du, et~al.
  2021{\natexlab{a}}.
\newblock Few-shot learning with multilingual language models.
\newblock \emph{arXiv preprint arXiv:2112.10668}.

\bibitem[{Lin et~al.(2021{\natexlab{b}})Lin, Liu, Moon, Crook, Zhou, Wang, Yu,
  Madotto, Cho, and Subba}]{lin2021leveraging}
Zhaojiang Lin, Bing Liu, Seungwhan Moon, Paul~A Crook, Zhenpeng Zhou, Zhiguang
  Wang, Zhou Yu, Andrea Madotto, Eunjoon Cho, and Rajen Subba.
  2021{\natexlab{b}}.
\newblock Leveraging slot descriptions for zero-shot cross-domain dialogue
  statetracking.
\newblock In \emph{Proceedings of the 2021 Conference of the North American
  Chapter of the Association for Computational Linguistics: Human Language
  Technologies}, pages 5640--5648.

\bibitem[{Litschko et~al.(2021)Litschko, Vuli{\'c}, Ponzetto, and
  Glava{\v{s}}}]{litschko2021evaluating}
Robert Litschko, Ivan Vuli{\'c}, Simone~Paolo Ponzetto, and Goran Glava{\v{s}}.
  2021.
\newblock Evaluating multilingual text encoders for unsupervised cross-lingual
  retrieval.
\newblock In \emph{European Conference on Information Retrieval}, pages
  342--358. Springer.

\bibitem[{Liu et~al.(2021{\natexlab{a}})Liu, Shen, Zhang, Dolan, Carin, and
  Chen}]{liu2021makes}
Jiachang Liu, Dinghan Shen, Yizhe Zhang, Bill Dolan, Lawrence Carin, and Weizhu
  Chen. 2021{\natexlab{a}}.
\newblock What makes good in-context examples for gpt-$3 $?
\newblock \emph{arXiv preprint arXiv:2101.06804}.

\bibitem[{Liu et~al.(2021{\natexlab{b}})Liu, Yuan, Fu, Jiang, Hayashi, and
  Neubig}]{liu2021pre}
Pengfei Liu, Weizhe Yuan, Jinlan Fu, Zhengbao Jiang, Hiroaki Hayashi, and
  Graham Neubig. 2021{\natexlab{b}}.
\newblock Pre-train, prompt, and predict: A systematic survey of prompting
  methods in natural language processing.
\newblock \emph{arXiv preprint arXiv:2107.13586}.

\bibitem[{Liu et~al.(2020)Liu, Gu, Goyal, Li, Edunov, Ghazvininejad, Lewis, and
  Zettlemoyer}]{liu2020multilingual}
Yinhan Liu, Jiatao Gu, Naman Goyal, Xian Li, Sergey Edunov, Marjan
  Ghazvininejad, Mike Lewis, and Luke Zettlemoyer. 2020.
\newblock Multilingual denoising pre-training for neural machine translation.
\newblock \emph{Transactions of the Association for Computational Linguistics},
  8:726--742.

\bibitem[{Louvan and Magnini(2020)}]{louvan2020recent}
Samuel Louvan and Bernardo Magnini. 2020.
\newblock Recent neural methods on slot filling and intent classification for
  task-oriented dialogue systems: A survey.
\newblock In \emph{Proceedings of the 28th International Conference on
  Computational Linguistics}, pages 480--496.

\bibitem[{Madotto et~al.(2021)Madotto, Lin, Winata, and Fung}]{madotto2021few}
Andrea Madotto, Zhaojiang Lin, Genta~Indra Winata, and Pascale Fung. 2021.
\newblock Few-shot bot: Prompt-based learning for dialogue systems.
\newblock \emph{arXiv preprint arXiv:2110.08118}.

\bibitem[{Madotto et~al.(2020)Madotto, Liu, Lin, and
  Fung}]{madotto2020language}
Andrea Madotto, Zihan Liu, Zhaojiang Lin, and Pascale Fung. 2020.
\newblock Language models as few-shot learner for task-oriented dialogue
  systems.
\newblock \emph{arXiv preprint arXiv:2008.06239}.

\bibitem[{Perez et~al.(2021)Perez, Kiela, and Cho}]{perez2021true}
Ethan Perez, Douwe Kiela, and Kyunghyun Cho. 2021.
\newblock True few-shot learning with language models.
\newblock \emph{arXiv preprint arXiv:2105.11447}.

\bibitem[{Petroni et~al.(2019)Petroni, Rockt{\"a}schel, Riedel, Lewis, Bakhtin,
  Wu, and Miller}]{petroni2019language}
Fabio Petroni, Tim Rockt{\"a}schel, Sebastian Riedel, Patrick Lewis, Anton
  Bakhtin, Yuxiang Wu, and Alexander Miller. 2019.
\newblock Language models as knowledge bases?
\newblock In \emph{Proceedings of the 2019 Conference on Empirical Methods in
  Natural Language Processing and the 9th International Joint Conference on
  Natural Language Processing (EMNLP-IJCNLP)}, pages 2463--2473.

\bibitem[{Radford et~al.(2019)Radford, Wu, Child, Luan, Amodei, and
  Sutskever}]{Radford2019LanguageMA}
Alec Radford, Jeff Wu, Rewon Child, David Luan, Dario Amodei, and Ilya
  Sutskever. 2019.
\newblock Language models are unsupervised multitask learners.

\bibitem[{Raffel et~al.(2020)Raffel, Shazeer, Roberts, Lee, Narang, Matena,
  Zhou, Li, and Liu}]{raffel2020exploring}
Colin Raffel, Noam Shazeer, Adam Roberts, Katherine Lee, Sharan Narang, Michael
  Matena, Yanqi Zhou, Wei Li, and Peter~J Liu. 2020.
\newblock Exploring the limits of transfer learning with a unified text-to-text
  transformer.
\newblock \emph{Journal of Machine Learning Research}, 21:1--67.

\bibitem[{Scao et~al.(2022)Scao, Fan, Akiki, Pavlick, Ili{\'c}, Hesslow,
  Castagn{\'e}, Luccioni, Yvon, Gall{\'e} et~al.}]{scao2022bloom}
Teven~Le Scao, Angela Fan, Christopher Akiki, Ellie Pavlick, Suzana Ili{\'c},
  Daniel Hesslow, Roman Castagn{\'e}, Alexandra~Sasha Luccioni, Fran{\c{c}}ois
  Yvon, Matthias Gall{\'e}, et~al. 2022.
\newblock Bloom: A 176b-parameter open-access multilingual language model.
\newblock \emph{arXiv preprint arXiv:2211.05100}.

\bibitem[{Schick and Sch{\"u}tze(2021{\natexlab{a}})}]{schick2021few}
Timo Schick and Hinrich Sch{\"u}tze. 2021{\natexlab{a}}.
\newblock Few-shot text generation with natural language instructions.
\newblock In \emph{Proceedings of the 2021 Conference on Empirical Methods in
  Natural Language Processing}, pages 390--402.

\bibitem[{Schick and Sch{\"u}tze(2021{\natexlab{b}})}]{schick2021s}
Timo Schick and Hinrich Sch{\"u}tze. 2021{\natexlab{b}}.
\newblock It’s not just size that matters: Small language models are also
  few-shot learners.
\newblock In \emph{Proceedings of the 2021 Conference of the North American
  Chapter of the Association for Computational Linguistics: Human Language
  Technologies}, pages 2339--2352.

\bibitem[{Schuster et~al.(2019)Schuster, Gupta, Shah, and
  Lewis}]{schuster-etal-2019-cross-lingual}
Sebastian Schuster, Sonal Gupta, Rushin Shah, and Mike Lewis. 2019.
\newblock \href {https://doi.org/10.18653/v1/N19-1380} {Cross-lingual transfer
  learning for multilingual task oriented dialog}.
\newblock In \emph{Proceedings of the 2019 Conference of the North {A}merican
  Chapter of the Association for Computational Linguistics: Human Language
  Technologies, Volume 1 (Long and Short Papers)}, pages 3795--3805,
  Minneapolis, Minnesota. Association for Computational Linguistics.

\bibitem[{Srivastava et~al.(2022)Srivastava, Rastogi, Rao, Shoeb, Abid, Fisch,
  Brown, Santoro, Gupta, Garriga-Alonso et~al.}]{srivastava2022beyond}
Aarohi Srivastava, Abhinav Rastogi, Abhishek Rao, Abu Awal~Md Shoeb, Abubakar
  Abid, Adam Fisch, Adam~R Brown, Adam Santoro, Aditya Gupta, Adri{\`a}
  Garriga-Alonso, et~al. 2022.
\newblock Beyond the imitation game: Quantifying and extrapolating the
  capabilities of language models.
\newblock \emph{arXiv preprint arXiv:2206.04615}.

\bibitem[{Timkey and van Schijndel(2021)}]{timkey2021all}
William Timkey and Marten van Schijndel. 2021.
\newblock All bark and no bite: Rogue dimensions in transformer language models
  obscure representational quality.
\newblock \emph{arXiv preprint arXiv:2109.04404}.

\bibitem[{Utama et~al.(2021)Utama, Moosavi, Sanh, and
  Gurevych}]{utama2021avoiding}
Prasetya Utama, Nafise~Sadat Moosavi, Victor Sanh, and Iryna Gurevych. 2021.
\newblock Avoiding inference heuristics in few-shot prompt-based finetuning.
\newblock In \emph{Proceedings of the 2021 Conference on Empirical Methods in
  Natural Language Processing}, pages 9063--9074.

\bibitem[{Wang et~al.(2018)Wang, Singh, Michael, Hill, Levy, and
  Bowman}]{wang2018glue}
Alex Wang, Amanpreet Singh, Julian Michael, Felix Hill, Omer Levy, and Samuel
  Bowman. 2018.
\newblock Glue: A multi-task benchmark and analysis platform for natural
  language understanding.
\newblock In \emph{Proceedings of the 2018 EMNLP Workshop BlackboxNLP:
  Analyzing and Interpreting Neural Networks for NLP}, pages 353--355.

\bibitem[{Wang(2021)}]{mesh-transformer-jax}
Ben Wang. 2021.
\newblock {Mesh-Transformer-JAX: Model-Parallel Implementation of Transformer
  Language Model with JAX}.
\newblock \url{https://github.com/kingoflolz/mesh-transformer-jax}.

\bibitem[{Winata et~al.(2022)Winata, Wu, Kulkarni, Solorio, and
  Preo{\c{t}}iuc-Pietro}]{winata2022cross}
Genta Winata, Shijie Wu, Mayank Kulkarni, Thamar Solorio, and Daniel
  Preo{\c{t}}iuc-Pietro. 2022.
\newblock Cross-lingual few-shot learning on unseen languages.
\newblock In \emph{Proceedings of the 2nd Conference of the Asia-Pacific
  Chapter of the Association for Computational Linguistics and the 12th
  International Joint Conference on Natural Language Processing}, pages
  777--791.

\bibitem[{Winata et~al.(2021)Winata, Madotto, Lin, Liu, Yosinski, and
  Fung}]{winata2021language}
Genta~Indra Winata, Andrea Madotto, Zhaojiang Lin, Rosanne Liu, Jason Yosinski,
  and Pascale Fung. 2021.
\newblock Language models are few-shot multilingual learners.
\newblock In \emph{Proceedings of the 1st Workshop on Multilingual
  Representation Learning}, pages 1--15.

\bibitem[{Xue et~al.(2021)Xue, Constant, Roberts, Kale, Al-Rfou, Siddhant,
  Barua, and Raffel}]{xue2021mt5}
Linting Xue, Noah Constant, Adam Roberts, Mihir Kale, Rami Al-Rfou, Aditya
  Siddhant, Aditya Barua, and Colin Raffel. 2021.
\newblock mt5: A massively multilingual pre-trained text-to-text transformer.
\newblock In \emph{Proceedings of the 2021 Conference of the North American
  Chapter of the Association for Computational Linguistics: Human Language
  Technologies}, pages 483--498.

\bibitem[{Zhang et~al.(2022)Zhang, Roller, Goyal, Artetxe, Chen, Chen, Dewan,
  Diab, Li, Lin et~al.}]{zhang2022opt}
Susan Zhang, Stephen Roller, Naman Goyal, Mikel Artetxe, Moya Chen, Shuohui
  Chen, Christopher Dewan, Mona Diab, Xian Li, Xi~Victoria Lin, et~al. 2022.
\newblock Opt: Open pre-trained transformer language models.
\newblock \emph{arXiv preprint arXiv:2205.01068}.

\bibitem[{Zhao et~al.(2021)Zhao, Wallace, Feng, Klein, and
  Singh}]{zhao2021calibrate}
Tony~Z Zhao, Eric Wallace, Shi Feng, Dan Klein, and Sameer Singh. 2021.
\newblock Calibrate before use: Improving few-shot performance of language
  models.
\newblock \emph{arXiv preprint arXiv:2102.09690}.

\end{thebibliography}
\bibliographystyle{acl_natbib}

\appendix

\section{Retrieved Examples}
\label{sec:retrieved-examples}

We show the retrieved nearest neighbor train samples with their distance relative to the test samples in Table~\ref{tab:examples-nearest-trec},
Table~\ref{tab:examples-nearest-snips}, Table~\ref{tab:examples-nearest-sst2}, Table~\ref{tab:examples-nearest-multi-nlu-en-es}, Table~\ref{tab:examples-nearest-mtop-en-es} for TREC, SNIPS, SST2, Multi-NLU, and MTOP, respectively.

\begin{table}[!ht]
    \centering
       \resizebox{\linewidth}{!}{
    \begin{tabular}{@{}ll@{}}
    \toprule
        \textbf{Test sample:} & What continent is Argentina on ? \\ 
        \textbf{Label:} & LOC \\
    \midrule
        \textbf{Nearest samples} \\
    \midrule 
        \textbf{Train 1:} & What continent is Bolivia on
? \\
        \textbf{Label:} & LOC \\
        \textbf{Distance:} & 0.2664 \\
    \midrule 
        \textbf{Train 2:} & What is spaceball played on ? \\
        \textbf{Label:} & DESC \\
        \textbf{Distance:} & 0.8701 \\
    \midrule 
        \textbf{Train 3:} & What country is Mount Everest in ? \\
        \textbf{Label:} & LOC \\
        \textbf{Distance:} & 0.8916 \\
    \midrule 
        \textbf{Train 4:} & What is the highest continent ? \\
        \textbf{Label:} & LOC \\
        \textbf{Distance:} & 0.8971 \\
    \midrule 
        \textbf{Train 5:} & How big is Australia ? \\
        \textbf{Label:} & NUM \\
        \textbf{Distance:} & 0.9021 \\
    \midrule 
        \textbf{Train 6:} & What is fiber
in food ?  \\
        \textbf{Label:} & DESC \\
        \textbf{Distance:} & 0.9265 \\
    \bottomrule
    \end{tabular}
      }
    \caption{Examples of Nearest Samples on TREC dataset.}
    \label{tab:examples-nearest-trec}
\end{table}

\begin{table}[!ht]
    \centering
       \resizebox{0.8\linewidth}{!}{
    \begin{tabular}{@{}ll@{}}
    \toprule
        \textbf{Test sample:} & find movie times \\ 
        \textbf{Label:} & search\_screening\_event \\
    \midrule
        \textbf{Nearest samples} \\
    \midrule 
        \textbf{Train 1:} & fine movie times \\
        \textbf{Label:} & search\_screening\_event  \\
        \textbf{Distance:} & 0.9162 \\
    \midrule 
        \textbf{Train 2:} & show movie times \\
        \textbf{Label:} & search\_screening\_event \\
        \textbf{Distance:} & 0.9412 \\
    \midrule 
        \textbf{Train 3:} & what movie schedule \\
        \textbf{Label:} & search\_screening\_event  \\
        \textbf{Distance:} & 1.0439 \\
    \midrule 
        \textbf{Train 4:} & list movie schedule \\
        \textbf{Label:} & search\_screening\_event \\
        \textbf{Distance:} & 1.0549 \\
    \midrule 
        \textbf{Train 5:} & find the movie curl \\
        \textbf{Label:} & search\_creative\_work \\
        \textbf{Distance:} & 1.1642 \\
    \midrule 
        \textbf{Train 6:} & find google news  \\
        \textbf{Label:} & search\_creative\_work  \\
        \textbf{Distance:} & 1.177859 \\
    \bottomrule
    \end{tabular}
      }
    \caption{Examples of Nearest Samples on SNIPS dataset.}
    \label{tab:examples-nearest-snips}
\end{table}

\begin{table}[!ht]
    \centering
       \resizebox{\linewidth}{!}{
    \begin{tabular}{@{}ll@{}}
    \toprule
        \textbf{Test sample:} & romanek keeps the film constantly taut ... \\
        & reflecting the character 's  \\ 
        & instability with a metaphorical visual style and \\
        & an unnerving , heartbeat-like score . , \\ 
        \textbf{Label:} & positive \\
    \midrule
        \textbf{Nearest samples} \\
    \midrule 
        \textbf{Train 1:} & despite the film 's bizarre developments , \\ 
        & hoffman keeps us riveted with every painful   \\
        & nuance , unexpected flashes of dark comedy and  \\
        & the character 's gripping humanity . \\
        \textbf{Label:} & positive \\
        \textbf{Distance:} & 0.6650 \\
    \midrule 
        \textbf{Train 2:} & perversely undercuts the joie de vivre \\ 
        & even as he creates it ,  giving the movie \\
        & a mournful undercurrent that places the   \\
        & good-time shenanigans in welcome perspective . \\
        \textbf{Label:} & positive \\
        \textbf{Distance:} & 0.6676 \\
    \midrule 
        \textbf{Train 3:} & hoffman keeps us riveted with every painful \\
        & nuance ,unexpected flashes of dark comedy and the  \\
        & character 's gripping humanity . \\
        \textbf{Label:} & positive \\
        \textbf{Distance:} & 0.6891 \\
    \midrule 
        \textbf{Train 4:} & the film occasionally tries the viewer 's patience \\
        & with slow pacing and a main character who   \\
        & sometimes defies sympathy, but it ultimately satisfies  \\
        & with its moving story . \\
        \textbf{Label:} & positive \\
        \textbf{Distance:} & 0.6894 \\
    \midrule 
        \textbf{Train 5:} & though nijinsky 's words grow increasingly \\
        & disturbed , the film maintains a beguiling  serenity  \\
        & and poise that make it accessible for a non-narrative \\
        & feature . \\
        \textbf{Label:} & positive \\
        \textbf{Distance:} & 0.7006 \\
    \midrule 
        \textbf{Train 6:} & with a tone as variable as the cinematography ,\\
        & schaeffer 's film never settles into .  \\
        & the light-footed enchantment the material needs ,  \\
        & and the characters ' quirks and foibles never jell \\
        & into charm  \\
        \textbf{Label:} & negative \\
        \textbf{Distance:} & 0.7034 \\
    \bottomrule
    \end{tabular}
      }
    \caption{Examples of Nearest Samples on SST2 dataset.}
    \label{tab:examples-nearest-sst2}
\end{table}

\begin{table}[!ht]
    \centering
       \resizebox{\linewidth}{!}{
    \begin{tabular}{@{}ll@{}}
    \toprule
        \textbf{Test sample:} & Cambia mi alarma de 5pm a 7pm esta noche. \\ 
        \textbf{Label:} & alarm/modify\_alarm \\
    \midrule
        \textbf{Nearest samples} \\
    \midrule 
        \textbf{Train 1:} & Change my alarm from 5pm to 7pm tonight. \\
        \textbf{Label:} & alarm/modify\_alarm  \\
        \textbf{Distance:} & 0.9318 \\
    \midrule 
        \textbf{Train 2:} & Change my alarm from 7am to 6:45am. \\
        \textbf{Label:} & alarm/modify\_alarm \\
        \textbf{Distance:} & 1.0793 \\
    \midrule 
        \textbf{Train 3:} & Change my alarm from 7am to 8am. \\
        \textbf{Label:} & alarm/modify\_alarm  \\
        \textbf{Distance:} & 1.1009 \\
    \midrule 
        \textbf{Train 4:} & Change my alarm tomorrow morning from \\
        & 7 am to 9 am. \\
        \textbf{Label:} & alarm/modify\_alarm \\
        \textbf{Distance:} & 1.1360 \\
    \midrule 
        \textbf{Train 5:} & Change my alarm from 4:45 am to 5:00 am \\
        & every weekday. \\
        \textbf{Label:} & alarm/modify\_alarm \\
        \textbf{Distance:} & 1.1441 \\
    \midrule 
        \textbf{Train 6:} & remind me to set my alarm for 8am \\
        & tomorrow morning at 10pm  \\
        \textbf{Label:} & reminder/set\_reminder  \\
        \textbf{Distance:} & 1.1559 \\
    \bottomrule
    \end{tabular}
      }
    \caption{Examples of Nearest Samples on Multi-NLU (en-es) dataset.}
    \label{tab:examples-nearest-multi-nlu-en-es}
\end{table}

\begin{table}[!ht]
    \centering
       \resizebox{0.87\linewidth}{!}{
    \begin{tabular}{@{}ll@{}}
    \toprule
        \textbf{Test sample:} & arrête la minuterie \\ 
        \textbf{Label:} & delete\_timer \\
    \midrule
        \textbf{Nearest samples} \\
    \midrule 
        \textbf{Train 1:} & discontinue timer \\
        \textbf{Label:} & delete\_timer  \\
        \textbf{Distance:} & 1.1686 \\
    \midrule 
        \textbf{Train 2:} & Pause the egg timer \\
        \textbf{Label:} & pause\_timer \\
        \textbf{Distance:} & 1.2166 \\
    \midrule 
        \textbf{Train 3:} & Stop the new timer \\
        \textbf{Label:} & pause\_timer  \\
        \textbf{Distance:} & 1.2331 \\
    \midrule 
        \textbf{Train 4:} & stop the timer \\
        & 7 am to 9 am. \\
        \textbf{Label:} & pause\_timer \\
        \textbf{Distance:} & 1.2342 \\
    \midrule 
        \textbf{Train 5:} & Start the stopwatch \\
        & every weekday. \\
        \textbf{Label:} & create\_timer \\
        \textbf{Distance:} & 1.2346 \\
    \midrule 
        \textbf{Train 6:} & mute the countdown please \\
        & tomorrow morning at 10pm  \\
        \textbf{Label:} & pause\_timer  \\
        \textbf{Distance:} & 1.2471 \\
    \bottomrule
    \end{tabular}
      }
    \caption{Examples of Nearest Samples on MTOP (en-es) dataset.}
    \label{tab:examples-nearest-mtop-en-es}
\end{table}

\section{Data Overlaps}
\label{sec:overlap}
We show the data overlaps on Table~\ref{overlap}.

\section{Data and Tasks Statistics}
We show the data and tasks information on Table~\ref{fig:task}.

\begin{table*}[!h]
\centering
\resizebox{\textwidth}{!}{
\begin{tabular}{lllll}
\toprule
\multirow{1}{*}{\textbf{Task}} & \multicolumn{1}{c}{\textbf{Dataset}} & \textbf{Lang(s)} &  \multicolumn{1}{c}{\textbf{Instruction}} & \multicolumn{1}{c}{\textbf{\#Labels}} \\ \midrule
\multirow{3}{*}{Intent Detection} & SNIPS~\cite{coucke2018snips} & en & \multirow{3}{*}{classify an intent from an utterance} & \multicolumn{1}{c}{3}\\
& Multi-NLU~\cite{schuster-etal-2019-cross-lingual} & en, es & & \multicolumn{1}{c}{12} \\
& MTOP~\cite{li-etal-2021-mtop} & en, es, de, fr & & \multicolumn{1}{c}{3-27} \\ \midrule
Question Classification & TREC~\cite{li-roth-2002-learning} & en & classify a label from a question & \multicolumn{1}{c}{6} \\ \midrule
Sentiment & SST-2~\cite{wang2018glue} & en & classify a sentiment from an utterance & \multicolumn{1}{c}{2} \\ \bottomrule
\end{tabular}
}
\caption{Tasks and Datasets.}
\label{fig:task}
\end{table*}

\begin{table}[!h]
\centering
\resizebox{0.75\linewidth}{!}{
\begin{tabular}{lr}
\toprule
\textbf{Task} & \textbf{Overlap Rate (\%)} \\ \midrule
\textbf{SNIPS} & 1.61 \\ \midrule
\textbf{Multi-NLU} \\
English (en) & 31.08 \\
Spanish (es) &  13.41 \\ \midrule
\textbf{MTOP} \\
English (en) & 1.00 \\
French (fr) & 1.71 \\
German (de) & 1.90 \\
Spanish (es) & 2.86 \\ \midrule
\textbf{TREC} & 1.49 \\ 
\midrule
\textbf{SST2} & 0.56 \\ 
\bottomrule
\end{tabular}
}
\caption{Sample Overlap Rate.}
\label{overlap}
\end{table}




\end{document}